\documentclass{article}

\usepackage{arxiv}
\usepackage[utf8]{inputenc}
\usepackage[T1]{fontenc}
\usepackage{hyperref}
\usepackage{url}
\usepackage{booktabs}
\usepackage{amsfonts}
\usepackage{nicefrac}
\usepackage{microtype}
\usepackage{lipsum}
\usepackage{graphicx}
\usepackage{natbib}
\usepackage{doi}

\usepackage{amsmath,amssymb,amsfonts}
\usepackage{textcomp}
\usepackage{algorithm}
\usepackage{algpseudocode}
\usepackage{booktabs}
\usepackage{subfig}
\usepackage{balance}
\usepackage{hyperref}
\usepackage{soul}

\usepackage{amsthm}
\usepackage{float}
\newtheorem{assumption}{Assumption}
\newtheorem{hypothesis}{Hypothesis}
\usepackage{placeins}
\usepackage[table]{xcolor} 
\usepackage{upgreek}

\newcommand{\fbseries}{\unskip\setBold\aftergroup\unsetBold\aftergroup\ignorespaces}

\newcommand{\setBoldness}[1]{\def\fake@bold{#1}}

\def\BibTeX{{\rm B\kern-.05em{\sc i\kern-.025em b}\kern-.08em
    T\kern-.1667em\lower.7ex\hbox{E}\kern-.125emX}}

\title{Image Score: Learning and Evaluating Human Preferences for Mercari Search}

\author{
  Chingis Oinar \\
  Mercari \\
  Tokyo, Japan \\
  \texttt{cowana@mercari.com} \\
  \And
  Miao Cao \\
  Mercari \\
  Tokyo, Japan \\ 
 \texttt{miao@mercari.com} \\
  \And
  Shanshan Fu \\
  Mercari \\
  Tokyo, Japan \\
  \texttt{s-fu@mercari.com}
}

\begin{document}
\maketitle

\begin{abstract}
Mercari is the largest C2C e-commerce marketplace in Japan, having more than 20 million active monthly users. Search being the fundamental way to discover desired items, we have always had a substantial amount of data with implicit feedback. Although we actively take advantage of that to provide the best service for our users, the correlation of implicit feedback for such tasks as image quality assessment is not trivial. Many traditional lines of research in Machine Learning (ML) are similarly motivated by the insatiable appetite of Deep Learning (DL) models for well-labelled training data. Weak supervision is about leveraging higher-level and/or noisier supervision over unlabeled data. Large Language Models (LLMs) are being actively studied and used for data labelling tasks. We present how we leverage a Chain-of-Thought (CoT) to enable LLM to produce image aesthetics labels that correlate well with human behavior in e-commerce settings. Leveraging LLMs is more cost-effective compared to explicit human judgment, while significantly improving the explainability of deep image quality evaluation which is highly important for customer journey optimization at Mercari. We propose a cost-efficient LLM-driven approach for assessing and predicting image quality in e-commerce settings, which is very convenient for proof-of-concept testing. We show that our LLM-produced labels correlate with user behavior on Mercari. Finally, we show our results from an online experimentation, where we achieved a significant growth in sales on the web platform. 
\end{abstract}


\section{Introduction}

\begin{figure}[t!]
\centering
\includegraphics[width=5.5cm]{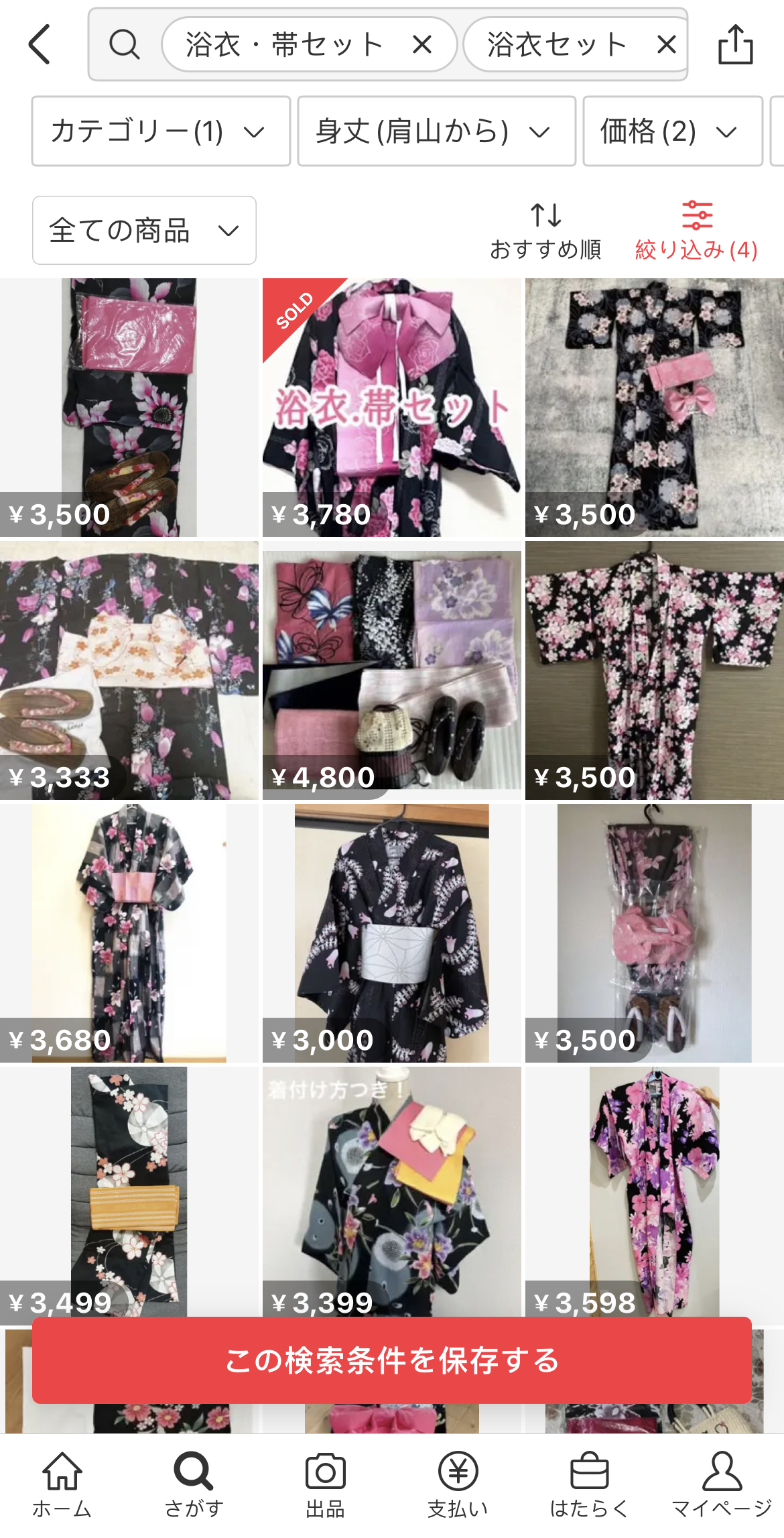}
\caption{The search result page user interface in the Mercari app. The figure shows how we display items to users, in a grid layout. The users only see images and prices at first. We study the importance of image quality when looking at similarly relevant and similarly priced items.} 
\label{fig:mercari}
\vspace{-10pt}
\end{figure}

Mercari is a two-sided online marketplace where anyone can buy and sell a diverse range of new and used products. A seller lists a unique item and a buyer starts searching for a product generally with a search query, probably specifying a product type or name. As the buyer explores and refines the query, the presented items typically tend to become more and more similar. This may be because multiple sellers sell the same product or a single seller creates multiple listings of the product. As Fig.~\ref{fig:mercari} shows, users expect to see images of items with the prices on the bottom left; hence, they usually do not see titles or any other information before tapping on the images. Therefore, it is valid to assume that images play a crucial role on Mercari in judging whether presented items are relevant to the query. The impact of image attributes on customer behavior may be even more significant when all items are relevant and similarly priced, since C2C marketplaces tend to have images of various qualities \citep{8658403}. Thus, an important part of building the desired search experience for our users is identifying how visual information affects user behavior. 

Mercari is committed to providing the best experience to both buyers and sellers. Sellers should be able to list their items easily: they can take a picture with their phone, provide a title, get suggestions on meta-data for the listing, and start attracting buyers. However, it is not trivial to provide high-quality photos without proper training or sufficient experience in the marketplace. On the other end, if the marketplace has an excessive amount of poor-quality product images, buyers might get an unsatisfactory experience. Therefore, we propose a cost-efficient LLM-driven approach for assessing and predicting image quality in e-commerce settings, which is very convenient for proof-of-concept testing as well.

Image quality evaluation has always been a very practical and important task in a wide variety of applications, including image ranking \citep{chang2017aesthetic, chen2023x} and synthetic image generation \citep{xu2024imagereward}. Although there are studies that demonstrate that the quality of images affects user behaviour on peer-to-peer marketplaces like Mercari, they also involve collecting explicit human judgements before exploring human preferences, which might consist of multiple stages of explicit annotation processes \citep{8658403}. Thus, these intermediate steps can be both time- and resource-consuming; hence, we aim to explore a more cost-efficient approach to assess the impact of images on user behaviour through implicit feedback and synthetic labels generated by LLMs.  

Our main contributions are summarized as follows:
\begin{itemize}
    \item We propose a data generation approach to decouple implicit feedback from relevance, thereby highlighting the contribution of image quality towards user clicks.
    \item We demonstrate that our LLM-produced image quality labels correlate with user behavior on Mercari, leveraging implicit feedback. Thus, we enable the use of clicks in offline model evaluation as one of the key metrics. 
    \item We show that the proposed Image Score model significantly outperforms relevance-based and popularity-based baseline models, such as predicting by CLIP score and historical click-through rate (CTR), at click prediction. This demonstrates that we successfully decouple relevance from implicit feedback. Moreover, Image Score achieved statistically significant growth in sales—almost a 7\% increase in terms of ATPU—on the Web platform, proving its potential for customer journey optimization on Mercari.
\end{itemize}

Our findings can be valuable for designing a cost-efficient image quality evaluation pipeline for marketplaces and analyzing the correlation between image quality and implicit feedback.
\section{Related Work}

\subsection{Weak Supervision via LLMs}
Weak supervision is about leveraging higher-level and/or noisier sources of training signals in an ML model. These sources could include heuristics, constraints, or predictions from expert models. Pang et al. propose ranking LLM responses to provide a better dialogue experience \citep{pang2023leveraging}. The ranking model is trained using labels generated by multiple smaller expert models, combined with heuristics; they propose factors such as the sentiment of the subsequent human response and its length. This approach makes the concept of an improved dialogue experience more understandable. Motivated by the cost-effectiveness and interpretability of weakly supervised learning, we explore the use of LLMs as the sole labeling function, introducing heuristics during the CoT. Sun et al. demonstrate that LLMs can produce high-quality labels, significantly outperforming models trained on human labels \citep{chatgpt@ranking}. Chen et al. propose using LLMs as an image quality judge for synthetically generated images, demonstrating a high correlation with human preference \citep{chen2023x}. They propose a CoT that covers multiple image quality attributes, guiding the LLM to score images out of 10 with these criteria in mind. Inspired by the ranking capability of LLMs and the interpretability introduced by CoT prompting, we explore the use of LLMs as an image quality judge in the marketplace domain. We also investigate the correlation of LLM-produced labels with human preference through implicit feedback, making our approach highly cost-effective, scalable, and convenient for proof-of-concept A/B testing. 

\subsection{Image Quality and Aesthetics Evaluation}
Early works on image aesthetics evaluation focus on handcrafted features \citep{bhattacharya2010framework, datta2006studying, li2010towards}. However, they require substantial manual engineering, which can be expensive in real-world business settings. Subsequently, with the rapid adoption of deep learning techniques, numerous supervised methods have shown superior performance \citep{pan2022vcrnet, cheon2021perceptual}. Nonetheless, they traditionally depend on human-labeled data, limiting their generalizability and versatility. Consequently, efforts have been made to explore zero-shot or weakly supervised approaches \citep{clipiqa, chen2023x}. CLIP-IQA demonstrates that pre-trained CLIP embeddings not only capture the content of images but also their quality attributes \citep{clipiqa}. We believe this finding holds significant value for real-world applications, where the reusability of pre-trained models can reduce costs and expedite proof-of-concept (PoC) experimentation. 

\subsection{Image Quality Evaluation in Online Marketplaces}
Belém et al. provide experiments on Elo7, the largest Brazilian marketplace, to analyze how the aesthetics of product images influence user interest, measured by the number of clicks \citep{Belm2019ImageAA}. The reported results confirm a correlation, which is particularly strong in specific product categories. Zakrewsky et al. explore the importance of visual characteristics on Etsy, focusing on how image features—including color, simplicity, scene composition, texture, style, aesthetics, and overall quality—influence purchase decisions \citep{DBLP:journals/corr/ZakrewskyAS16}. 

In the area of advertisement display, Azimi et al. leverage large-scale data from the RightMedia ads exchange system and identify a subset of high-impact visual features for predicting CTR \citep{azimi2012impact}. 
\section{Image Score: Learning and Evaluating Human Preferences for
Mercari Search}
\subsection{Data Collection and Label Generation}

\begin{figure}[t!]
\centering
\includegraphics[width=12.5cm]{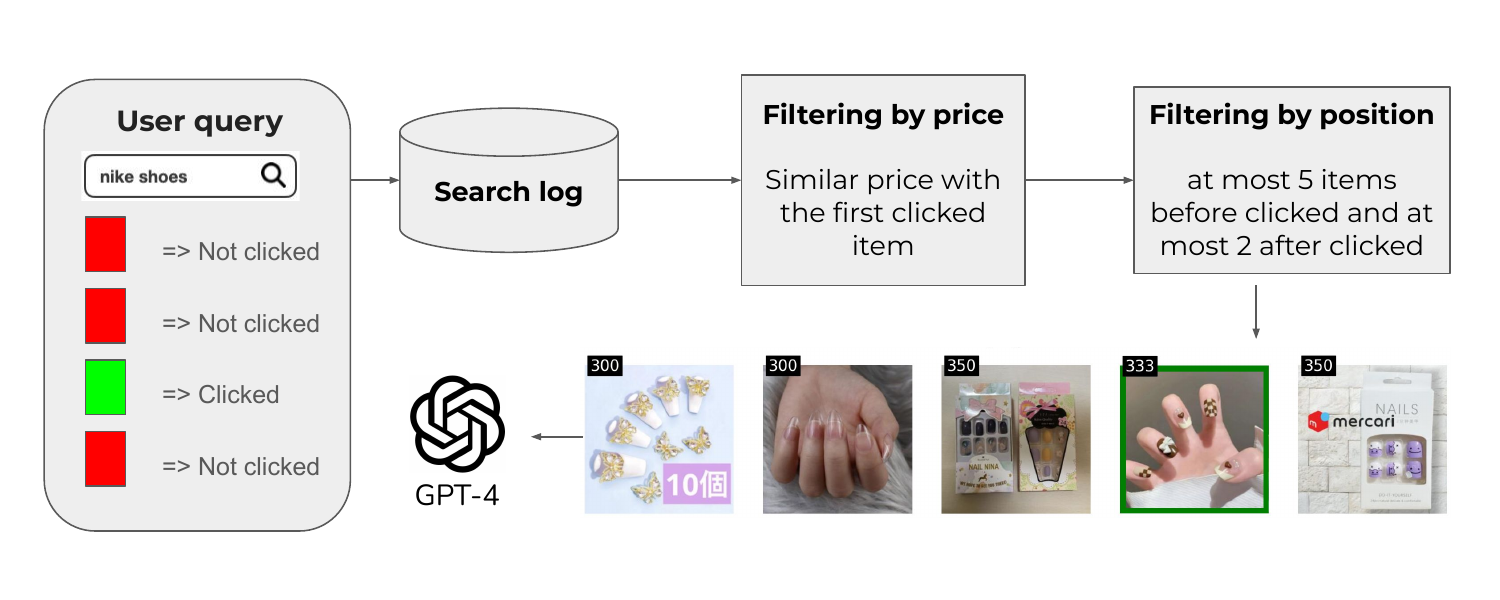}
\caption{The data collection and processing pipeline. Price filtering and position windowing are applied to SERPs to get similar item pairs.} 
\label{fig:data-collection}
\end{figure}

In this section, we describe our procedure for collecting batches of images from search logs, generating image quality scores utilizing LLMs, and assessing the correlation between the generated scores and user behavior on Mercari.

\subsubsection{Data Collection}
We performed data collection based on the following hypothesis:

\begin{hypothesis}
If two items are similarly relevant and similarly priced, the one with better image quality is likely to be more appealing, hence attracting more clicks.
\label{hypothesis:main}
\end{hypothesis}

Subsequently, we extracted the search logs within a specific date range and applied price filtering and position windowing to the original search results pages (SERPs), as shown in Fig.~\ref{fig:data-collection}. We randomly sampled training data consisting of 150{,}000 SERPs and validation data consisting of 36{,}104 SERPs.

To focus on items with similar prices, we filtered items based on their price difference from the first clicked item. We then limited the items by their position, retaining at most five items before the clicked item and up to two items after the clicked item, with only one clicked item per SERP. Thus, we extend the assumptions proposed by Li et al.~\citep{li2020handling}, which are motivated by the issue of position bias, and rely on the following two structural assumptions during data collection:

\begin{assumption}
The clicked item is observed and examined. The items above the clicked item are more likely to be observed and examined as well. 
\label{assumption:observation}
\end{assumption}

\begin{assumption}
Within some window, items ranked close to each other tend to be similarly relevant and similarly desirable given a context. 
\label{assumption:relevance}
\end{assumption}

The processed images from the same SERP were then batch-analyzed using GPT-4 through a vision API to achieve a relative quality evaluation \citep{achiam2023gpt}.

\subsubsection{Annotation with LLM}

\begin{figure*}[t!]
\centering
\includegraphics[width=6in]{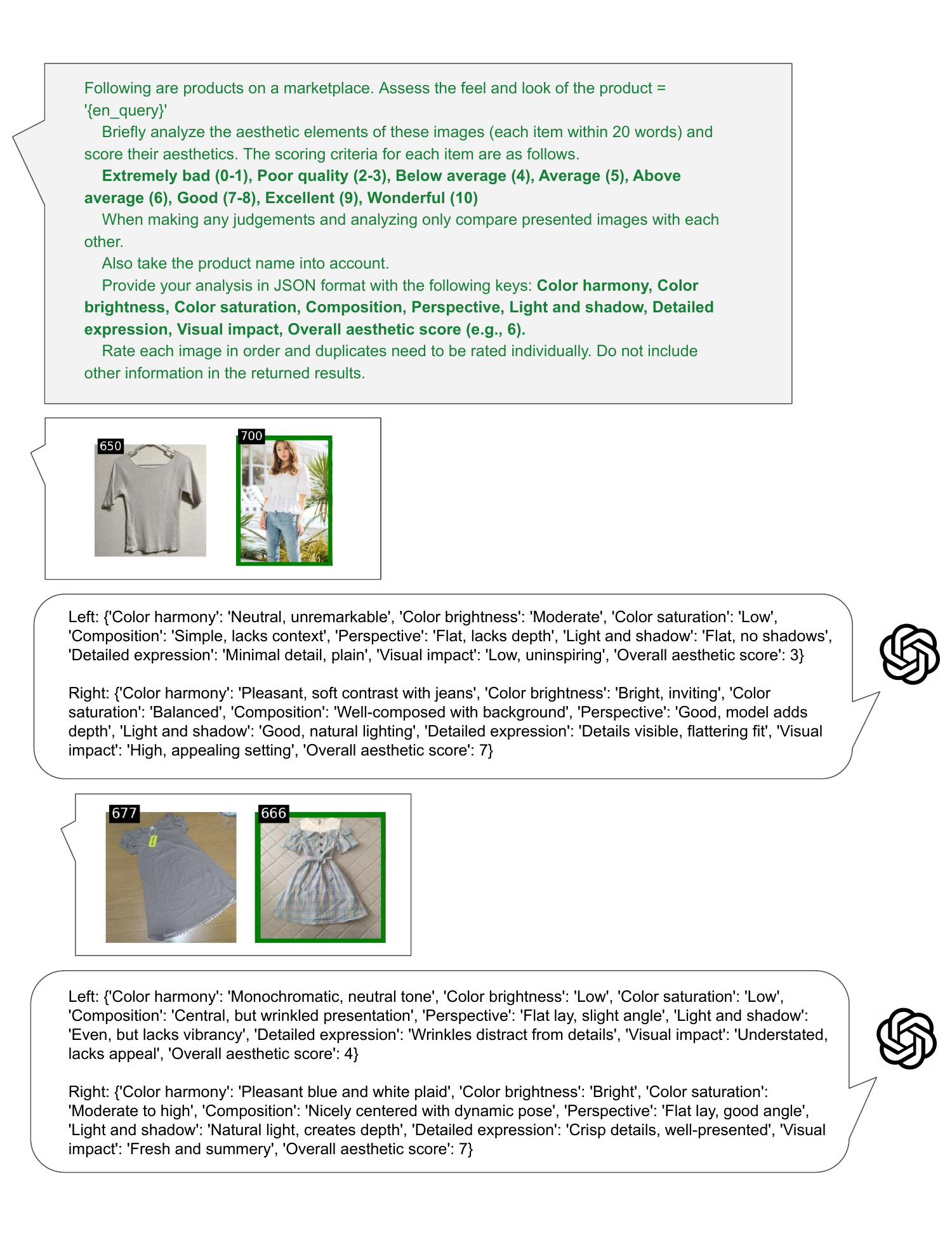}
\caption{The prompt for image aesthetic evaluation and image batch examples. The images with green borders are clicked items and others are not clicked. $en\_query$ denotes a SERP query translated into English.} 
\label{fig:prompt}
\vspace{-10pt}
\end{figure*}

Recent studies have shown that LLMs can produce high-quality labels for ranking applications \citep{chatgpt@ranking}. This means LLMs can identify a particular context and produce appropriate rankings based on the relevance of each document to the given task. For our data annotation, we employed LLMs to generate labels, leveraging the prompts developed by X-IQE~\citep{chen2023x} for the aesthetic evaluation stage. Since our goal is to establish a baseline for future iterations, we adopt the aesthetic evaluation procedure, as it has been shown to produce both highly consistent and high-quality labels that correlate with human preference~\citep{chen2023x}. However, we are also aware of the potential limitations of the proposed prompting strategy; more detailed, especially in relation to the e-commerce setting, Chains-of-Thought (CoT) prompting may be more effective \citep{cot}. Figure~\ref{fig:prompt} illustrates the prompt used and the corresponding model response, demonstrating the quality and interpretability of this approach.

\subsubsection{Dataset Analysis}

\begin{figure}[t!]
\centering
\includegraphics[width=8.5cm]{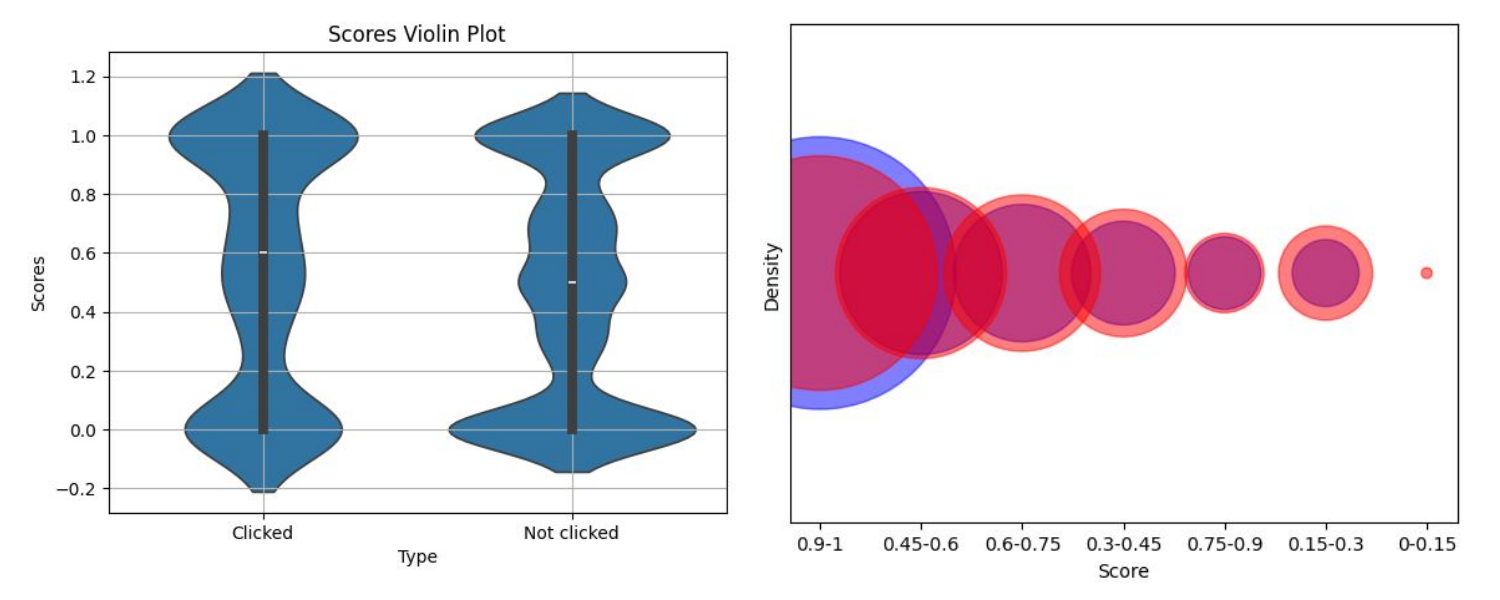}
\caption{The score distributions for clicked and non-clicked items. In the chart on the right-hand side, the colors blue and red represent clicked and non-clicked items, respectively.} 
\label{fig:score-distribution}
\end{figure}

To assess the correlation between LLM-produced image scores and user clicks, we compare the image scores for clicked and non-clicked items annotated by the LLM. The score distribution, where scores are normalized between 0 and 1 for each SERP, is shown in Fig.~\ref{fig:score-distribution}. As observed, clicked items generally receive higher image scores from the LLM, as both the median and average values are higher by a noticeable margin: 0.6 versus 0.5 for the median, and 0.5428 versus 0.4751 for the mean. Likewise, the Kolmogorov–Smirnov (K–S) test reveals that the differences in the distributions of LLM-produced annotations between clicked and non-clicked items are statistically significant (p-value: 8.367e-09).

\textbf{Discussion.} Additionally, we performed a deeper analysis comparing liked and purchased items. We found that the differences in distributions between liked items and non-clicked items, as well as between purchased items and non-clicked items, are statistically significant. However, this is not the case when comparing them with clicked items. This may indicate that image quality is not a dominant factor driving higher levels of engagement on Mercari (e.g., comments, likes, and purchases).

\begin{figure*}[t!]
\centering
\includegraphics[width=7in]{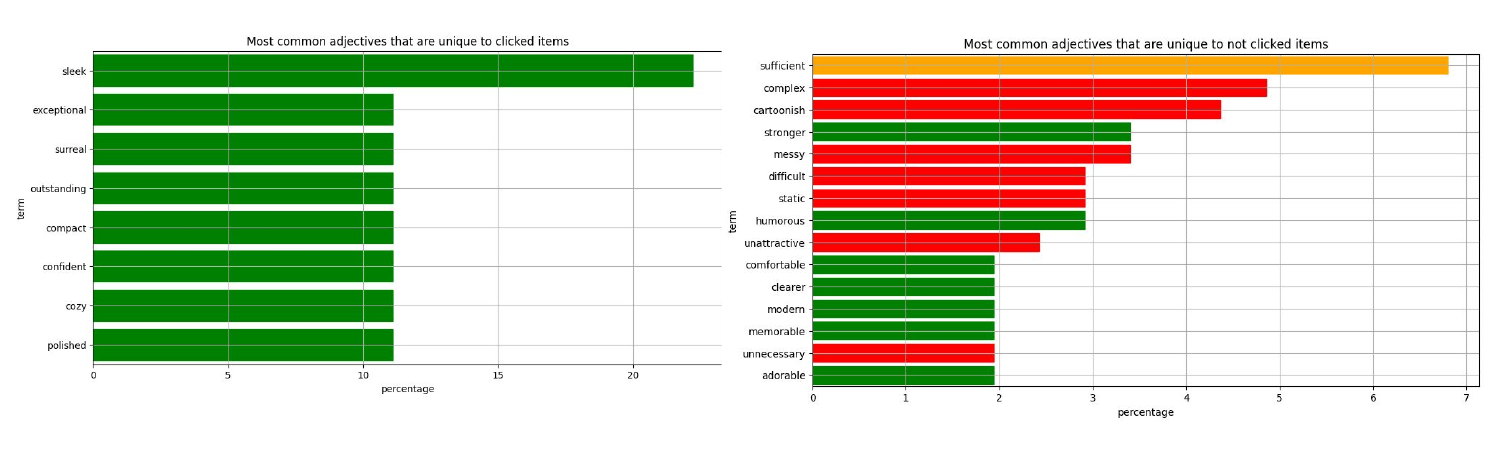}
\caption{Most common adjectives from LLM analysis that are unique to clicked and non-clicked items on the left and right sides, respectively. The green bar represents positive adjectives, the red bar represents negative adjectives, and the yellow bar represents neutral adjectives.} 
\label{fig:llm-reasoning}
\vspace{-10pt}
\end{figure*}

We compared the most common adjectives from LLM analysis responses for clicked and non-clicked items, as shown in Fig.~\ref{fig:llm-reasoning}. Non-clicked items received more negative evaluations from the LLM, which helps explain their lower scores. We provide a high-level overview in Appendix~\ref{appendix:B}, showing that despite the general predominance of positive adjectives, non-clicked items are more frequently associated with negative or neutral descriptors.

\subsection{Proposed Model}

Admittedly, performing inference with LLMs is ideal for validating our findings online; however, it is limited by inference costs and time constraints. Therefore, we aim to use LLM annotations to train a lighter ML model for online experiments, which is also convenient and cost-efficient for proof-of-concept (PoC) testing.

\begin{figure}[t!]
\centering
\includegraphics[width=8.5cm]{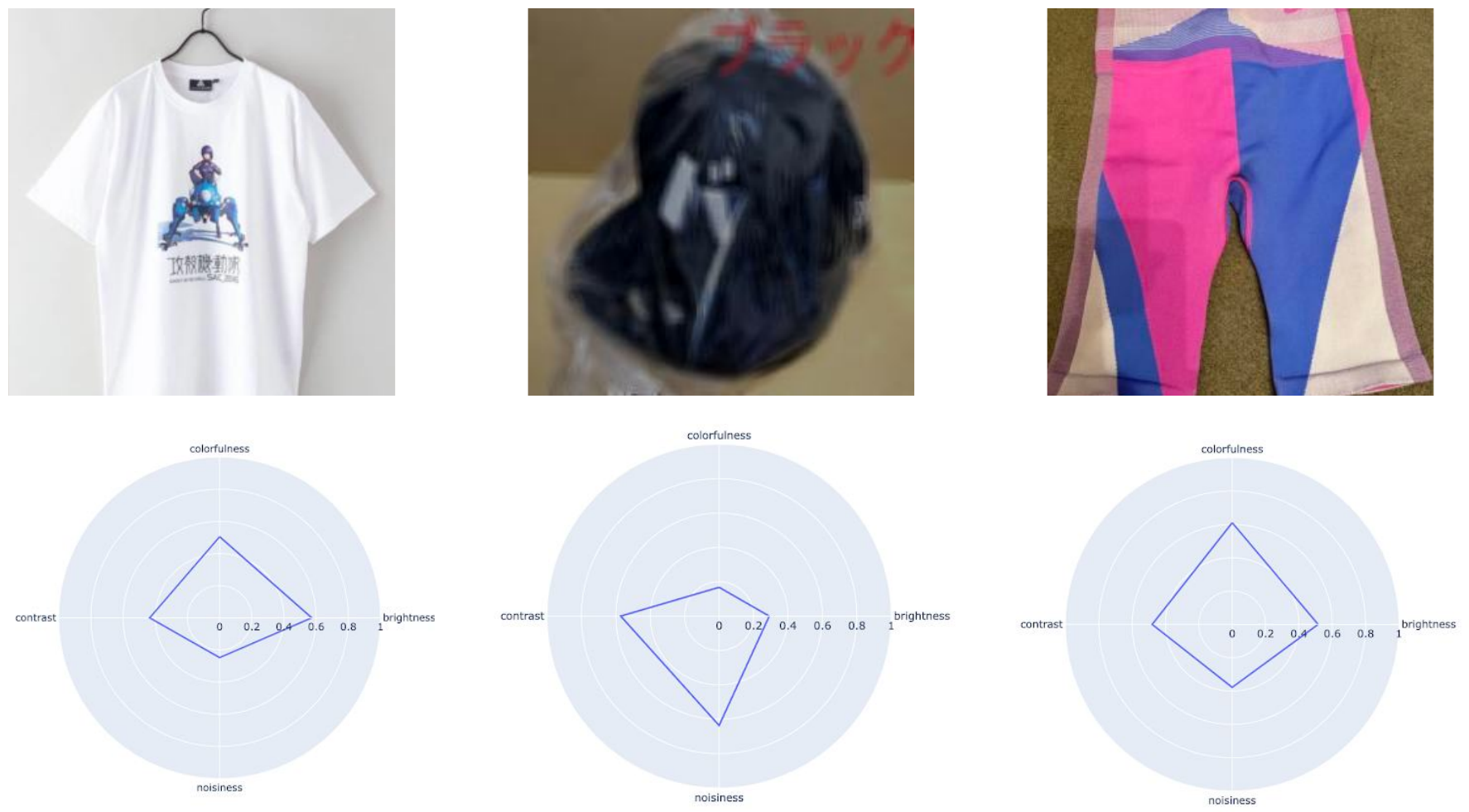}
\caption{We adopt the findings of visual perception proposed by CLIP-IQA. We find that CLIP embeddings are also applicable for assessing image quality on Mercari. Likewise, the assessment of colorfulness, brightness, noisiness, and contrast is performed without explicit task-specific training, using the proposed antonym prompts instead.} 
\label{fig:lookandfeel}
\end{figure}

Given the success and rapidly increasing adoption of CLIP-like models \citep{clip, siglip, sun2023eva, blip} in various vision tasks, we likewise explore their use within our domain. Moreover, recent works such as CLIP-IQA \citep{clipiqa} show that CLIP is effective for deep image quality assessment tasks, even in a zero-shot setting, since the produced embeddings capture not only content but also image quality attributes to a satisfactory extent. This property is particularly useful for our task. Therefore, we qualitatively examined CLIP-IQA on Mercari images and found that frozen CLIP embeddings can be used as the backbone for our architecture, as demonstrated in Fig.~\ref{fig:lookandfeel}. Thus, we adopt pre-trained CLIP image embeddings to establish our baseline, as this approach is highly time- and resource-efficient and allows us to expedite our project timeline.

\subsubsection{Preference Learning}

\begin{figure*}[t!]
\centering
\includegraphics[width=5in]{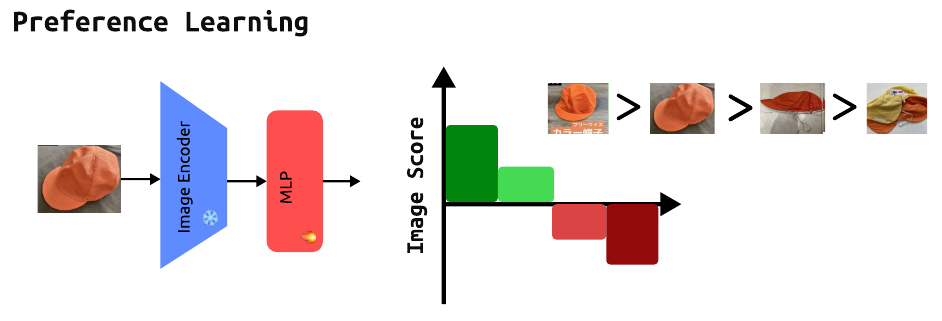}
\caption{We adopt the pre-trained image encoder of CLIP as the backbone of our architecture; hence, we only train an additional MLP layer. Preference learning is performed using Eqn.~\ref{eqn:loss}, where we encourage the MLP to rank $k$ (up to 8) images in the order produced by the LLM.} 
\label{fig:framework}
\vspace{-10pt}
\end{figure*}

In addition, we fine-tune a multi-layer perceptron (MLP) on top of the pre-trained CLIP embeddings to optimize specifically for our domain. This allows us to benefit from CLIP’s representation learning while maintaining task-specific adaptation.

Similar to the reward model (RM) training proposed by ImageReward \citep{xu2024imagereward}, we formulate our preference learning task as a pairwise learning-to-rank (LTR) problem. After obtaining $k$ images (up to 8) ranked by the LLM for a query $Q$, where the order from best to worst is denoted as $x_1, x_2, ..., x_k$, we construct up to $C^{2}_{k}$ pairs, assuming no ties between images. It is important to note that, besides clicked items, we only consider images that are scored lower by the LLM than the clicked ones. In doing so, we aim to learn both image quality attributes and domain-specific factors that drive user engagement.

Thus, for each pair $(x_i, x_j)$ where $x_i$ is ranked higher than $x_j$ by the LLM (i.e., $y_i > y_j$), the loss function is defined as:
\begin{equation}
L(x_i, x_j) = \sum_{i=1}^{N} \sum_{j=1}^{N} I[y_i > y_j] \cdot \text{FL}([f_{\theta}(x_{i}), f_{\theta}(x_{j})], y=0).
\label{eqn:loss}
\end{equation}
where $f_{\theta}(x_{i})$ and $f_{\theta}(x_{j})$ are predicted scores $s_i$ and $s_j$ from the MLP parameterized by $\theta$, and $\text{FL}$ denotes Focal Loss \citep{focalloss}.

\textbf{Discussion.} Given the noisy nature of our problem, we find that Focal Loss significantly improves model performance, as it prevents a large number of easy negatives from overwhelming the model during training. Similar to ImageReward, we observe rapid convergence followed by overfitting, where Focal Loss helps mitigate this issue. Additionally, we perform a careful grid search over hyperparameters based on the validation set to further control overfitting.

\subsection{Deployment}

Our search system is based on the Elasticsearch (ES) search engine~\citep{elasticsearch}. We propose incorporating image quality scores into Elasticsearch to influence search ranking. This section details how we serve the Image Score model and index its predictions to integrate image quality scores into our existing indexing pipeline.
 
\subsubsection{Image Score Component} 
We leverage the Triton inference server~\citep{triton} to host the Image Score model. Our current data processing pipeline for indexing (a.k.a.\ analyzer) runs on Google Cloud Dataflow~\citep{dataflow}. To incorporate the Image Score model, we integrate an image scoring component into the Dataflow job, enabling it to communicate with the Triton inference server via a gRPC client.

\subsubsection{Offline Prediction} 
For historical items, or items that already exist in the search index, we run the offline indexing pipeline to include image scores. To optimize computational resources during the experimental phase, we limit this process to items created within the past 90 days.

\subsubsection{Online Prediction} 
\begin{figure*}[t!]
\centering
\includegraphics[width=5in]{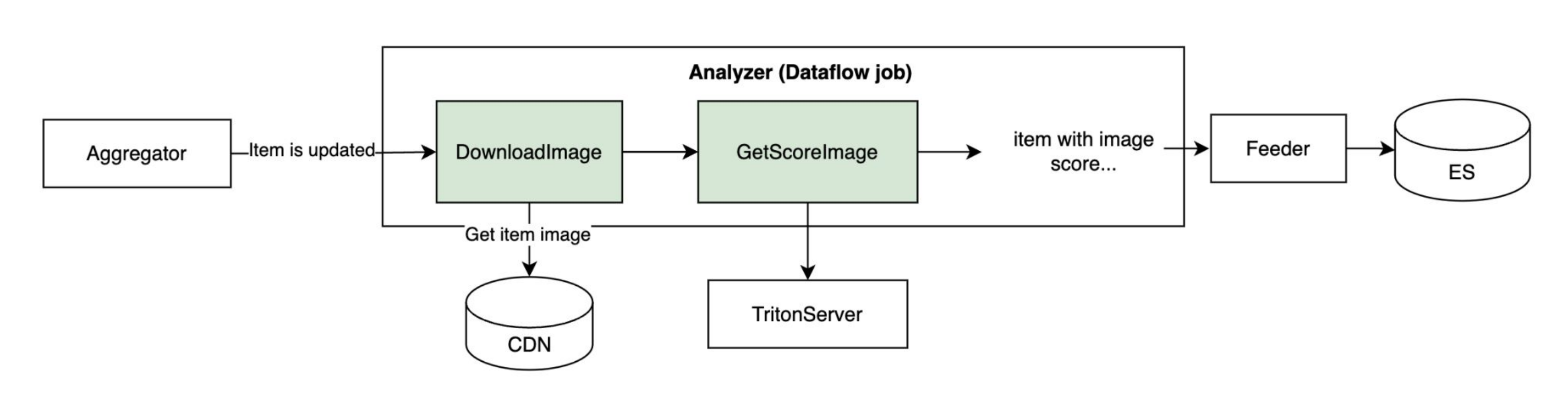}
\caption{The Elasticsearch online indexing pipeline with the Image Score model.} 
\label{fig:indexing}
\end{figure*}

Fig.~\ref{fig:indexing} illustrates the online indexing pipeline with the image scoring component. When an item is updated, a Pub/Sub message containing raw item information is sent from the aggregator to the analyzer. The image score is predicted synchronously within the analyzer, and the processed item, along with its image score, is then indexed via the feeder into the Elasticsearch index.

\subsection{Boosting in Retrieval Stage}

This section describes the procedure used to incorporate the image score into the existing Elasticsearch mechanism, with the goal of enhancing search performance.

\subsubsection{Boosting Formula}

The current ranking formula in our Elasticsearch implementation is based on a combination of features, including the relevance of an item to a search query and its relevance to associated metadata. To extend this formulation, we integrate image scores using both multiplicative and additive approaches, defined as follows:

\textbf{Multiplicative Approach:}
\begin{equation}
\text{final\_score} = \text{existing\_relevance\_score} \cdot \text{image\_score}.
\label{eqn:es_multiplicative}
\end{equation}

\textbf{Additive Approach:}
\begin{equation}
\text{final\_score} = \text{existing\_relevance\_score} + \text{image\_score}.
\label{eqn:es_additive}
\end{equation}

The multiplicative approach amplifies the impact of the image score in conjunction with other factors (e.g., relevance score, category boost score), emphasizing the importance of both relevance and high-quality images. In contrast, the additive approach incorporates the image score directly into the relevance component, ensuring a more balanced contribution of image quality in the final ranking. Note that $\text{existing\_relevance\_score}$ is used in subsequent computations to derive $\text{final\_score}$; therefore, the relative contribution of $\text{image\_score}$ may be attenuated in downstream stages.

The effectiveness of each approach—multiplicative or additive—is evaluated through rigorous online experiments to determine the most effective strategy for optimizing search performance and user experience.

\textbf{Discussion.} In this experiment, we use $\text{image\_score}$ solely to influence the Elasticsearch-produced ranking of documents. However, in practice, the item ranking produced by Elasticsearch is subsequently passed to a downstream machine learning ranking model. Therefore, modifications at the Elasticsearch stage can still influence the final re-ranked results.

\subsubsection{Challenges in Offline Evaluation}

A key challenge in offline evaluation arises from the dynamic nature of the freshness factor in our Elasticsearch implementation. This factor, which reflects the recency of an item, is an important component of the scoring function. However, extracting this dynamic component from Elasticsearch explanation outputs is non-trivial, making it difficult to accurately reconstruct scores for offline evaluation.

\section{Offline Experimental Results}

Offline experimentation is a crucial first step before deploying our ML model into production. Moreover, since our primary goal is to establish a baseline for future iterations, it is important to benchmark the performance of our initial model. In this section, we first introduce our training and evaluation setups, followed by a presentation of our offline and online results.

\subsection{Training Setting} 
For the image embedding network, we adopt ViT-B/16 pre-trained on the translated CC12M dataset \citep{sawada2024release, rinna-japanese-clip-vit-b-16}, where captions are in Japanese. We train models on NVIDIA Tesla T4 GPUs with a batch size of 128. As stated earlier, we keep the parameters of CLIP fixed and only train an additional MLP on top. The model is trained using the AdamW optimizer \citep{adamw}, with a learning rate of $1 \times 10^{-4}$ and a weight decay of $1 \times 10^{-3}$. We adopt the cosine annealing learning rate scheduler \citep{loshchilov2016sgdr} and train the model for 100 epochs. For Focal Loss \citep{focalloss}, we set the $\gamma$ parameter to 2.

\subsection{Evaluation Metrics} 

\textbf{Ordered Pair Accuracy (OPA).} Given a pair of images with different labels, OPA measures the probability that the model assigns a higher score to the image with the higher label. The metric is defined as:
\begin{equation}
OPA({y}, {s}) = \frac{\sum_{i=1}^{N} \sum_{j=1}^{N} I[s_i > s_j] \cdot I[y_i > y_j]}{\sum_{i=1}^{N} \sum_{j=1}^{N} I[y_i > y_j]}.
\label{eqn:opa}
\end{equation}
where $s_i$ and $y_i$ denote the predicted score and the target label (produced by the LLM) for image $x_i$, respectively.

\textbf{Click Accuracy (CA).} We construct image pairs where one image is strictly clicked, measuring the probability that the model assigns a higher score to the clicked image. This metric reflects how well the model predicts user engagement based on image quality:
\begin{equation}
CA({c}, {s}) = \frac{\sum_{i=1}^{N} \sum_{j=1}^{N} I[s_i > s_j] \cdot I[c_i = 1 \land c_j = 0]}{\sum_{i=1}^{N} \sum_{j=1}^{N} I[c_i = 1 \land c_j = 0]}.
\label{eqn:ca}
\end{equation}
where $s_i$ and $c_i$ denote the predicted score and the click label for image $x_i$, respectively.

\textbf{Discussion.} Due to potential noise in LLM-generated labels, CA provides a more reliable indicator of model performance in predicting user engagement, whereas OPA may be more sensitive to label noise. This comparison also highlights the contribution of image quality attributes to click prediction.

\begin{table}[t!]
\caption{Offline results (in \%) on a proprietary test dataset. CLIP Score and CTR Highest predictions are determined by the highest cosine similarity between an item title and a query, and by the highest historical click-through rate, respectively. We use CLIP pre-trained on translated CC12M \citep{sawada2024release, rinna-japanese-clip-vit-b-16}, where captions are in Japanese.}
\label{tb:proprietary}
\begin{center}
\begin{tabular}{ l|c|c} 
\toprule
Method & OPA & CA\\
\hline \hline
Random Guess & 50.0 & 50.0 \\
CTR Highest & 42.59 & 44.40 \\
CLIP Score & 49.75 & 50.20 \\
Image Score (CE) & 67.45 & 64.14 \\
Image Score (FL) & \textbf{67.75} & \textbf{64.42} \\
\bottomrule
\end{tabular}
\end{center}
\end{table}

\subsection{Results on a Proprietary Test Dataset} 

As shown in Table~\ref{tb:proprietary}, our model outperforms all baselines. Our best-performing variant, Image Score (FL), achieves 67.75\% in OPA and 64.42\% in CA, representing improvements of 17.75\% and 14.42\% over random guessing, respectively. Additionally, we observe that Focal Loss (FL) improves model performance over Cross Entropy (CE), consistent with findings from ImageReward \citep{xu2024imagereward}. We attribute this improvement to the robustness of FL to noisy labels during training.

Moreover, we observe that using historical CTR leads to worse performance than random guessing, with 42.59\% in OPA and 44.40\% in CA when predicting future clicks. Similarly, CLIP Score does not significantly outperform random guessing in click prediction, exceeding it by only 0.2\%. Revisiting Hypothesis~\ref{hypothesis:main}, we conclude that clicks are effectively decoupled from relevance in our dataset. Consequently, Image Score is able to achieve significantly better performance in click prediction.

In Appendix~\ref{appendix:A}, we provide qualitative examples comparing the original SERP ordering with the ordering produced by Image Score. As observed, clicked images (highlighted with green frames) tend to rank higher due to their superior image quality.
\section{Online Experimental Results}

\begin{table*}[t!]
    \centering
    \begin{tabular}{|l|c|c|c|c|} \hline 
        
        Metric Name & variant1/control & t-value (variant1 vs.\ control) & variant2/control & t-value (variant2 vs.\ control) \\ \hline  
        BCR & \textcolor{blue}{102.15\%} & \textcolor{blue}{1.9738} & 101.50\% & 1.3762 \\ \hline  
        BCR (via search) & 101.90\% & 1.5734 & 100.97\% & 0.8072 \\ \hline  
        ATPU & \textcolor{blue}{106.99\%} & \textcolor{blue}{2.22} & 100.75\% & 0.2560 \\ \hline 
        ATPU (via search) & \textcolor{blue}{106.66\%} & \textcolor{blue}{2.10} & 100.18\% & 0.0611 \\ \hline
        
    \end{tabular}
    \caption{A/B test results on the web platform. We performed a two-sided t-test, where variant1 and variant2 correspond to Eqns.~\ref{eqn:es_additive} and~\ref{eqn:es_multiplicative}, respectively. Values highlighted in blue indicate statistically significant improvements.}
    \label{tab:metrics_comparison_t_values}
\end{table*}

\subsection{Online Experimental Setting}

To evaluate the effectiveness of Eqns.~\ref{eqn:es_multiplicative} and~\ref{eqn:es_additive}, we conducted an online experiment by splitting users evenly into three groups: a baseline (control) group, an additive (variant1) group, and a multiplicative (variant2) group.

\subsection{A/B Test Results}

As shown in Table~\ref{tab:metrics_comparison_t_values}, the evaluation metrics reveal statistically significant improvements for users on the web platform. In particular, Average Transactions per User (ATPU) increased by 6.99\% in the marketplace. While we observe clear gains in user engagement and transaction activity for web users, mobile platforms exhibit negative impacts.

\textbf{Discussion.} The A/B test results suggest that web users respond positively to the modified Elasticsearch ranking, whereas mobile users exhibit different behavioral patterns. One possible explanation is that web users are more sensitive to visual quality due to larger image resolutions compared to mobile devices. This indicates that learning an optimal weighting of image quality within a downstream ML ranking model could be a promising direction for future improvements.

\subsection{Limitations and Future Work}

\subsubsection{High Scores for AI-Generated Images.}
Some users exploit the system by posting AI-generated images that receive high image quality scores. To mitigate this issue, it is necessary to introduce mechanisms that prevent such images from achieving high rankings solely based on visual quality. This may involve incorporating detection models for AI-generated content or refining prompting strategies to better account for authenticity. In this context, trustworthiness is as important as aesthetic quality in shaping user perception within the marketplace.

\subsubsection{Continuous Monitoring and Feedback Loop.}
To further improve transparency and robustness, it is important to establish a continuous monitoring system that tracks the performance of ranking models and user engagement metrics. Such a feedback loop would enable rapid iteration and more reliable evaluation of future ranking strategies.
\section{Conclusion}
In this study, we explore the impact of image quality on user behavior on the Mercari e-commerce app. First, we hypothesize that items with higher image quality lead to increased user engagement, clicks and transactions. Based on the hypothesis, we create an image quality assessment dataset labelled with LLM and train an ML image scoring model. We show the results from the online experiment to verify this hypothesis in a real-world setting. Our experiments reveal that incorporating image quality features significantly improves CTR and conversion rate.

Key findings from our study include:

\begin{enumerate}
\item Correlation Between Image Quality and User Behavior: There is a significant correlation between image quality in search results and user behavior, with higher quality images leading to increased user engagement, clicks and transactions.
\item Platform-Specific Image Preferences: Users may have different preferences or perceptions of images on different platforms, indicating that the contribution of image quality may vary depending on the platform used.
\item Limitations of Common Image Quality Metrics: While common image quality assessment metrics are somewhat correlated with user preferences, they do not fully capture the nuances of image quality in the e-commerce domain. This suggests a need for more tailored metrics that better reflect user preferences in this context.
\end{enumerate}

Our research contributes to the growing knowledge of visual perception in e-commerce and provides a foundation for future studies. Moving forward, we recommend a further exploration into behavioral factors based on the platform and the development of more sophisticated AI-generated content detection methods.
\section*{Acknowledgement}
We would like to express our sincere appreciation to our colleagues at Mercari for their contributions in making this project a reality, including Antoine Lecubin, Kentaro Takiguchi, Ryan Ginstrom and Pathompong Yupensuk.

\bibliographystyle{ACM-Reference-Format}
\bibliography{arxiv_version}

\section{APPENDICES}
\appendix
\section{Image Score: Qualitative Samples}
\label{appendix:A}
\begin{figure*}[t!]
\centering
\includegraphics[width=0.8\textwidth]{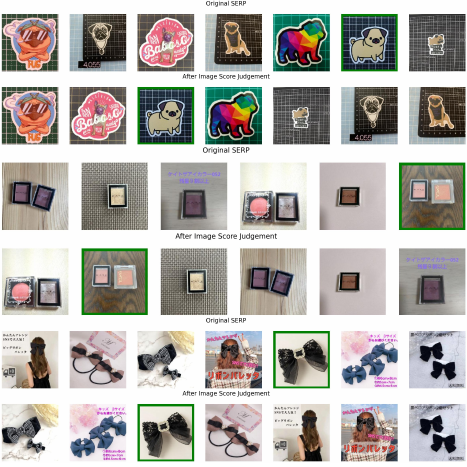}
\caption{We show the effect of the proposed Image Score model by comparing the original ordering in the Search Engine Results Page (SERP) with the ordering produced by the Image Score. The images annotated with green frames represent the items that have been clicked. The ranking in SERP is increasing from right to left, so the one on the left is displayed first.} 
\label{fig:qualitative_imagescore}
\end{figure*}
\FloatBarrier

\section{LLM Analysis: Most Common Adjectives in Clicked and Unclicked Items.}
\label{appendix:B}
\begin{figure*}[h!]
\centering
\includegraphics[width=1\textwidth]{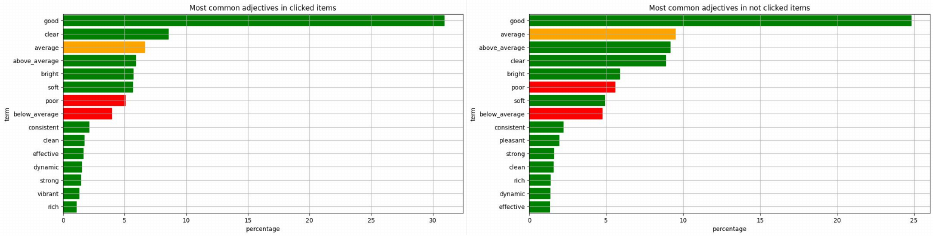}
\caption{We show the most common adjectives from LLM analysis for both clicked and not clicked items. As observed, the presence of positive adjectives generally dominate in both clicked and not clicked items. However, not clicked items tend to receive negative or neutral adjectives more frequently compared to clicked items.} 
\label{fig:qualitative_imagescore_example2}
\end{figure*}
\FloatBarrier

\end{document}